\documentclass[10pt]{asme2ej}
\usepackage{times}
\usepackage{url}
\usepackage{latexsym}
\usepackage{amsmath}
\usepackage[utf8]{inputenc} % allow utf-8 input
\usepackage[T1]{fontenc}    % use 8-bit T1 fonts
\usepackage{hyperref}       % hyperlinks
\usepackage{url}            % simple URL typesetting
\usepackage{booktabs}       % professional-quality tables
\usepackage{amsfonts}       % blackboard math symbols
\usepackage{nicefrac}       % compact symbols for 1/2, etc.
\usepackage{microtype}      % microtypography
\usepackage{multirow}       % multiple rows in table
\usepackage{amssymb}        % greek letters
\usepackage{graphicx}
\usepackage{makecell}

\title{DeepTitle - Leveraging BERT to generate \\Search Engine Optimized Headlines}

\author{Cristian Anastasiu\thanks{All authors contributed equally to this paper.}
    \affiliation{
	Amazon Web Services \\ Seefeldstrasse 69 \\ 8008 Z\"urich, Switzerland \\ canast@amazon.com    }	
}

\author{Hanna Behnke$^*$ 
    \affiliation{ SPRING Axel Springer Digital \\News Media GmbH \& Co. KG \\ Axel-Springer-Straße 65 \\10969 Berlin, Germany\\ hanna.behnke@spring-media.de
    }
}

\author{Sarah Lueck$^*$
    \affiliation{SPRING Axel Springer Digital \\ News Media GmbH \& Co. KG \\ Axel-Springer-Straße 65 \\10969 Berlin, Germany\\ sarah.lueck@spring-media.de
    }
}
\author{Viktor Malesevic$^*$
    \affiliation{Amazon Web Services \\ Oskar-von-Miller-Ring 20 \\ 80333 M\"unchen, Germany \\ malesv@amazon.com
    }
}
\author{Aamna Najmi$^*$
    \affiliation{Amazon Web Services \\ Oskar-von-Miller-Ring 20 \\ 80333 M\"unchen, Germany \\ anajmi@amazon.com
    }
}
\author{Javier Poveda-Panter$^*$
    \affiliation{Amazon Web Services \\ Oskar-von-Miller-Ring 20 \\ 80333 M\"unchen, Germany \\ jpovedap@amazon.com
    }
}

\begin{document}

\maketitle    

%%%%%%%%%%%%%%%%%%%%%%%%%%%%%%%%%%%%%%%%%%%%%%%%%%%%%%%%%%%%%%%%%%%%%%
\begin{abstract}
{Automated headline generation for online news articles is not a trivial task - machine generated titles need to be grammatically correct, informative, capture attention and generate search traffic without being "click baits" or "fake news". In this paper we showcase how a pre-trained language model can be leveraged to create an abstractive news headline generator for German language. We incorporate state of the art fine-tuning techniques for abstractive text summarization, i.e. we use different optimizers for the encoder and decoder where the former is pre-trained and the latter is trained from scratch. We modify the headline generation to incorporate frequently sought keywords relevant for search engine optimization. We conduct experiments on a German news data set and achieve a ROUGE-L-gram F-score of 40.02. Furthermore, we address the limitations of ROUGE for measuring the quality of text summarization by introducing a sentence similarity metric and human evaluation. 
}
\end{abstract}

\section{Introduction}
\label{intro}

\begin{figure}[!htbp]
      \centering
      \includegraphics[width=12cm]{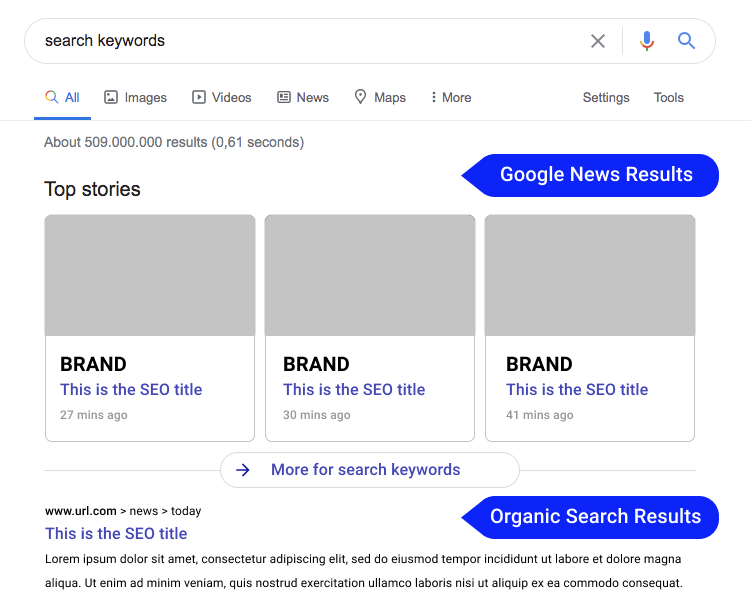}
      \caption{Visualization of Google search results that highlights the position of the SEO title in Google News boxes and organic search results.}
      \label{fig:google-search-results}
    \end{figure}

Most publishing houses generate a large share of revenue through digital products. Regardless of the business model, it is vital to increase the discoverability of the content to attract new customers to the website. Besides social media, search engines - especially Google - are the most important places for content discovery. To increase the likelihood of an article to be viewed by the user, the articles have to rank high up in the organic search results as well as in Google News boxes, cp. Fig. \ref{fig:google-search-results}. The lower the position in the result list, the lower the click through rate (CTR) for each search result \cite{ctr-google}. It is the core task of search engine optimization (SEO) to increase the visibility of the content and analyse which factors have a positive influence on organic search traffic. The article's title that is displayed in the search results is one of these factors \cite{Giomelakis2019} - hereinafter called "SEO title".

In this paper, we propose an approach to generate SEO titles based on the article's text using an encoder-decoder network. The project was conducted in collaboration with WELT, a German news media brand, which is part of Axel Springer SE. By recommending search engine optimized titles to the editors, we aim to accelerate the production of articles and increase organic search traffic. We use the expert knowledge of the WELT SEO team to build our model and use a manually labelled dataset to determine high-quality SEO titles. According to the experts, the article headline must catch the readers' attention and have a well-defined set of characteristics: most notably, SEO titles should be descriptive, intriguing as well as contain keywords that are specific to the context and frequently sought for. Additionally, the titles should not exceed the length of what Google displays in the organic search results.

%%%%%%%%%%%%%%%%%%%%%%%%%%%%%%%%%%%%%%%%%%%%%%%%%%%%%%%%%%%%%%%%%%%%%%
\section{Background}

\subsection{Abstractive Summarization}
In the editorial workflow, the SEO title is most commonly created after the article was written. In its core, SEO title generation is therefore the process of writing the shortest possible summary of the article. From a technical perspective, summarization can either be extractive or abstractive. While the former re-combines parts, usually sentences, from the original text \cite{extractive2001}, the latter creates new pieces of text to form a coherent, fluent output. The first neural, abstractive summarization models were introduced by \cite{rush2015abstractive} and \cite{nallapati2016abstractive}. Both used an encoder-decoder architecture to map a longer sequence of tokens to a shorter sequence, which became the de-facto standard for abstractive summarization. Given the strict limitation of the title's length and strong condensation of meaning, we choose abstractive summarization as the most suitable approach. 

Among the first to use the encoder-decoder architecture for title generation, \cite{lopyrev2015} trains a recurrent neural network with LSTM units and attention on news articles, using only the first 50 words of the article as input. In 2019, the term "extreme summarization" was introduced \cite{extremeSummarization}, an abstractive approach to create single-sentence summaries of documents using convolutional neural networks to capture long-range dependencies. \cite{vasilyev2019headline} proposes another interesting approach to title generation: A neural network is trained on 1.5 million article-title pairs, limiting the vocabulary used for the title prediction to those words that occur in the article to avoid grammatical and factual errors. Additionally, a double-blind trial approach is proposed as an alternative to traditional evaluation metrics like BLEU \cite{papineni-bleu} and ROUGE \cite{lin-rouge} to judge the quality of the generated title. 

\subsection{Transfer Learning for Summarization}
While these papers demonstrate that encoder-decoder architectures are well-suited for summarization tasks, the models are trained from scratch, requiring large quantities of training data. In recent years, it has been shown that transfer learning is a powerful technique to be applied in NLP.  Pre-trained models, such as ELMo \cite{elmo}, GPT \cite{gpt} and most prominently BERT \cite{bert_paper} are used to solve versatile NLP problems. BERT (Bidirectional Encoder Representations from Transformers) is bidirectionally trained on large quantities of text under the use of masking and next sentence prediction to model language context and, in its original form, applied to various NLP tasks like language understanding and translation. In 2019, \textsc{BertSumAbs} \cite{presumm} was published, a modification of the BERT architecture for the task of text summarization by incorporating interval segment embeddings. 

There are more recent end-to-end pre-trained models such as Pegasus \cite{zhang2019pegasus} and ProphetNet \cite{prophetnet} that do not need to be trained from scratch on the decoder side and have yet achieved state of the art results in abstractive summarization tasks on public datasets. Pegasus uses a standard transformer based encoder-decoder architecture with a novel self-supervised objective called gap sentence generation where whole sentences are masked instead of only tokens as it is the case in BERT. ProphetNet is also a transformer based encoder-decoder architecture that uses multiple stream self-attention mechanisms. 

We use \textsc{BertSumAbs} \cite{presumm} for SEO title generation because both Pegasus and ProphetNet are pre-trained on an English corpus only. The availability of a pre-trained German BERT \cite{germanbert} allows us to use it as a pre-trained encoder in the \textsc{BertSumAbs} architecture and apply it to German news articles by only training the decoder from scratch. In comparison, both Pegasus and ProphetNet would need to be entirely pre-trained on a massive German corpus before fine-tuning it on our news article dataset.

We further adapt the generation of abstractive summaries to meet our needs. We modify the length penalty to fulfill the needs of SEO titles and we introduce a rank penalty to favor summaries containing specific words.

%%%%%%%%%%%%%%%%%%%%%%%%%%%%%%%%%%%%%%%%%%%%%%%%%%%%%%%%%%%%%%%%%%%%%%
\section{Data and Preprocessing}
We train and evaluate our model on a data set from the German news brand WELT published on www.welt.de between 2014 and 2018. The article texts consist of the text itself and, in most of the cases, a brief introduction displayed in the beginning of the text. We only include articles from news-relevant departments. We filter out articles with a total word count above 512 and below 30 and SEO titles with a word count above 12 or below 3.

The cleaned data set consists of over 500,000 German news articles. Table \ref{table: trainset} presents how we split the data set.
In particular, we create two test sets for automated and human evaluation respectively. These test sets comprise SEO titles that were manually checked by SEO title experts (see Section \ref{sec:automaticeval} and \ref{sec:humaneval}).
\begin{table}[h]
  \caption{Data set statistics: size of training, validation, and test sets for automated and human evaluation and average document and summary length (in terms of words and sentences).}
  \label{table: trainset}
  \centering
  \begin{tabular}{c|cc|cc}
    \toprule
    \multirow{2}{*}{\# docs (train/val/test\textunderscore auto/test\textunderscore manual)} & \multicolumn{2}{c|}{avg. article length} & \multicolumn{2}{c}{avg. SEO title length} \\
     & words & sentences & words & sentences \\ 
    \midrule
    500,000/10,000/1000/100 & 167 & 12.2 & 6.5 & 1 \\
    \bottomrule
  \end{tabular}
\end{table}

%%%%%%%%%%%%%%%%%%%%%%%%%%%%%%%%%%%%%%%%%%%%%%%%%%%%%%%%%%%%%%%%%%%%%%
\section{Towards generating SEO Titles}
To create SEO titles which generate high search volumes, one needs two ingredients: a readable, informative title and keywords relating to the article with a high search volume. We address these needs in two approaches. First, we create a one-sentence-summary of the text by deploying a neural network with an encoder-decoder structure to generate news titles and propose a modified length penalty to meet the length requirements of an SEO title. Second, we generate a list of keywords ranked by their relevance to the text content and expected search volume in Google. In the end, we combine the two and use the list of keywords to tweak the generation of SEO titles by introducing a rank penalty.

\subsection{Model Architecture for Abstractive Summarization}
\label{marker}
BERT \cite{bert_paper} is a language model that learns contextual representations by being pre-trained with masked language modeling and next sentence prediction tasks.
This model can be fine-tuned to a specific custom natural language processing task. The BERT model outputs context rich vectors for each input token. However, as the original BERT architecture is limited to tokens and sentence pairs only, it is unable to gather sentence level representations. Thus, it is not suitable for text summarization. \cite{presumm} found one way to overcome this problem by using interval segment embeddings to distinguish different sentences. This way it is possible to extract sentence level representations in a document \cite{presumm}.
We use the work of \cite{presumm} to build an abstractive neural text summarization model, termed \textsc{BertSumAbs}. 
We encode the input article text as a sequence of continuous representations and decode it to generate a target summary token by token using a sequence to sequence encoder-decoder-architecture. We leverage the German BERT pre-trained language model and use it as the encoder to get context rich representations from our input. For the decoder, we use a 6 layered Transformer \cite{transformer} which is initialized randomly as indicated in \cite{presumm}. Since there is a mismatch between the encoder and decoder as the former is pre-trained whereas the latter is trained from scratch, we use the fine-tuning technique introduced in \cite{presumm}. This technique uses separate optimizers for the encoder and the decoder. We use a low learning rate for the encoder and a high learning rate for the decoder. This ensures that the encoder does not overfit and the decoder does not underfit.

\subsection{Tweaking the Generation of SEO Titles}

\label{gen_inst}
When generating SEO titles two important factors come in play: the length and the keywords contained in them. The title should have an optimal length to be descriptive and compact enough to fit in the limited pixel space of news search results. Additionally, if a title contains keywords with high search volume and relevance, it will more likely be found and rank well on a search engine.

To create SEO titles with these attributes, we modify Beam Search. Beam Search generates several candidate headlines (or beams) at inference. Each candidate is characterized by the probability score of all tokens. Only the top beams with highest scores serve as headlines. We use the idea introduced by \cite{gnmt} to modify the beam scores to incorporate the attributes. To do so, we penalize beams which either do not have the right length or do not contain keywords.

In general terms we modify for each beam the beam score $s(Y,X)$ with a penalty term $p(Y)$:
\begin{equation} \label{eq_len1}    
s(Y,X) = \frac{\log(P(Y|X))}{p(Y)},
\end{equation}
where $Y$ is the sequence of the beam, and $P(Y|X)$ the probability score of that beam given an input sequence $X$.

\subsubsection{Modified Length Penalty}
To meet the requirements of SEO titles with a optimal length $r$, in our case 12 tokens, we modify the length penalty introduced by \cite{gnmt}. The length penalty $lp$ then reads:

\begin{equation} \label{eq_len1}
lp(Y) = \frac{(5 + (\theta(Y, r) + 1))^\alpha}  {(5 + 1)^\alpha}
\end{equation}

with

\begin{equation} \label{eq_len1}    
\theta(Y, r) = 
\begin{cases}
    |Y| ,& \text{ if } |Y|  < r \\
    2r - |Y| , & \text{ otherwise}.
    \end{cases}
\end{equation}

\begin{figure}
     \centering
     \includegraphics[width=1\textwidth]{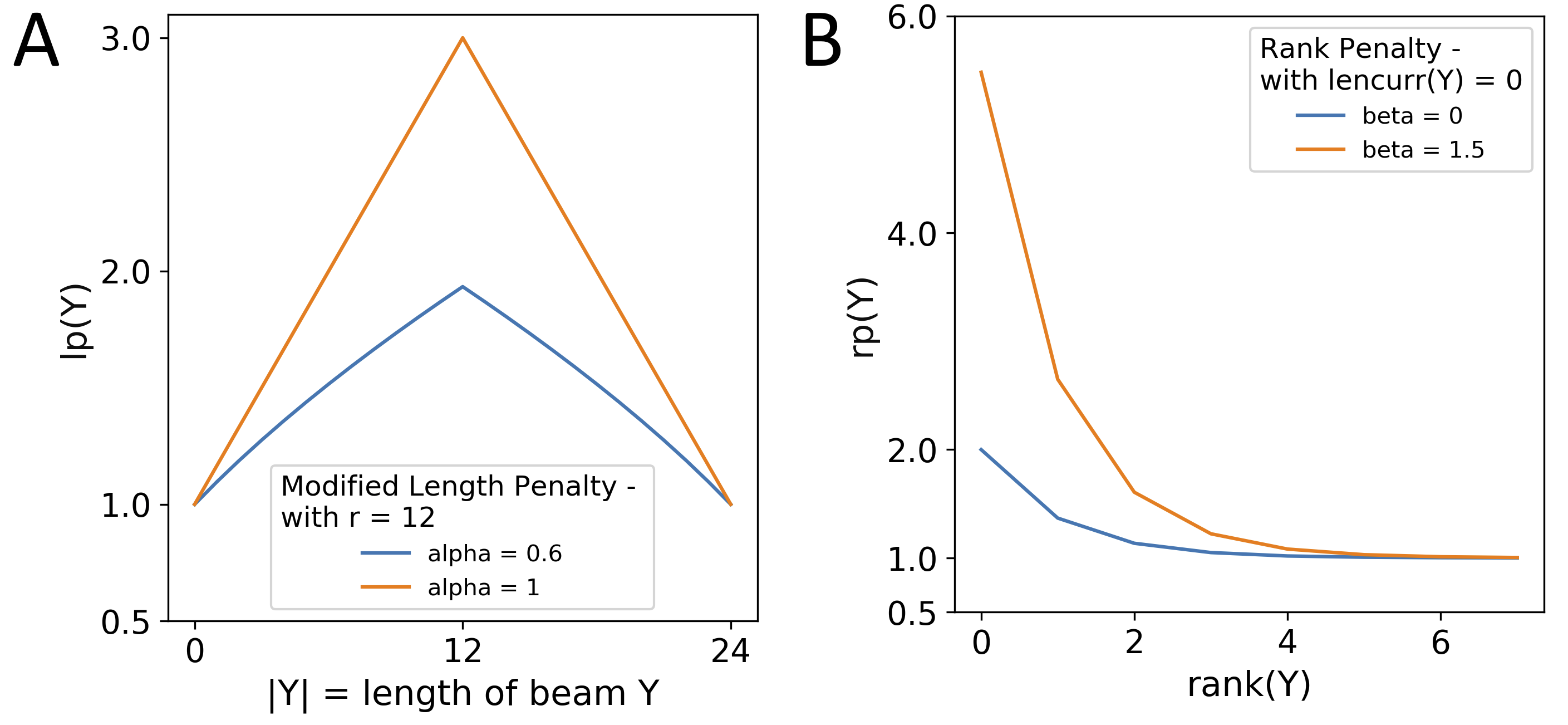}
    \caption{\textbf{Visualization of length and rank penalties.} \textbf{A} Modified length penalty $lp(Y)$: Every candidate headline with a length different from $r$ will be penalized. $\alpha$ defines the impact of the penalty, the higher $\alpha$ the higher the penalty influence. \textbf{B} Rank penalty $rp(Y)$: Candidate headlines containing keywords that ranked well (0, 1 or 2 for example) will have an improved score.}
     \label{fig:penalties}
 \end{figure}

$|Y|$ is the current candidate length and $\alpha$ is the length normalization coefficient as introduced in \cite{gnmt}. $\theta$ is a triangular function with its maximum at $r$, resulting in $lp(Y)$ following a similar triangular profile (cp. Fig.  \ref{fig:penalties} )

\subsubsection{Keyword Rank Penalty}

To include keywords into the generated SEO title we first create a list of keywords of the article ranked according to their relevance for the text and their search volume. To do this, we first use an off-the-shelf named entity recognition service to identify the relevant keywords in the article and capture their relevance score and syntactical properties. We gather the relative search volumes for all keywords in the article. As an additional feature, we also calculate the variation in search volume for the last three days, counting from the publication date of the news article. We add other corpus related features, namely tf-idf values and position of the keyword in the text. All these features act as inputs to an XGBoost ranking algorithm \cite{xgboost}. We train the model on a training set of 17,000 articles with titles manually generated or reviewed by search engine optimization experts. To create the training set of ranked keywords, we define the rank by the position of the keyword in the title, the first keyword is assigned the best (lowest) rank. If a keyword from the text is not found in the title it is not assigned any rank.

The keywords together with their rank, influence the generation of the SEO title as follows:
Similar to the length penalty, we create a rank penalty $rp$ which favors the beams which are generating words with a good (=low) rank.
Furthermore, the rank penalty is constructed to favor keywords at the beginning of the sentence, while the impact towards the end of the sentence is marginal. This way, we ensure that the rank penalty does not impact the grammatical correctness of the predicted sentences by altering words in the middle of the prediction. We transfer the keywords into subtokens. For every step in the sequence generation, we calculate the rank penalty for the predicted subtokens as follows:
\begin{equation} \label{eq_rank1}
rp(Y) = 1 + \exp\left(-\text{rank}(Y) - \frac{\text{len}_\text{curr}(Y)}{3} + \beta \right), 
\end{equation}
where rank is the rank from the corresponding keyword, $\text{len}_\text{curr}$ is the current length of the sequence $Y$. Dividing by 3 has been chosen and tested empirically in our use case but can be adapted. $\beta$ is a hyperparameter that we introduce to accentuate or dampen the impact of the rank penalty (cp. Fig. \ref{fig:penalties}).

As a consequence, if a beam contains a well ranking keyword at the beginning of the sentence, $rp$ will be high. If it contains a low ranking keyword, or the keyword is at the end of the sentence, $rp$ will be close to 1, almost not modifying the score of the beam Y.
If a beam does not contain a word from the keyword list, we set $rp$ to equal 1 by default.

%%%%%%%%%%%%%%%%%%%%%%%%%%%%%%%%%%%%%%%%%%%%%%%%%%%%%%%%%%%%%%%%%%%%%%
\section{Evaluation}

 To evaluate the quality of the generated titles, we employed both automated and human techniques. We used a data set of 1000 already published articles to generate SEO titles for automated evaluation and ot of these 100 for human evaluation. These SEO titles were compared the experts' manually created titles.

\subsection{Automated Evaluation}
 \label{sec:automaticeval}
\begin{figure}
     \centering
     \includegraphics[width=1\textwidth]{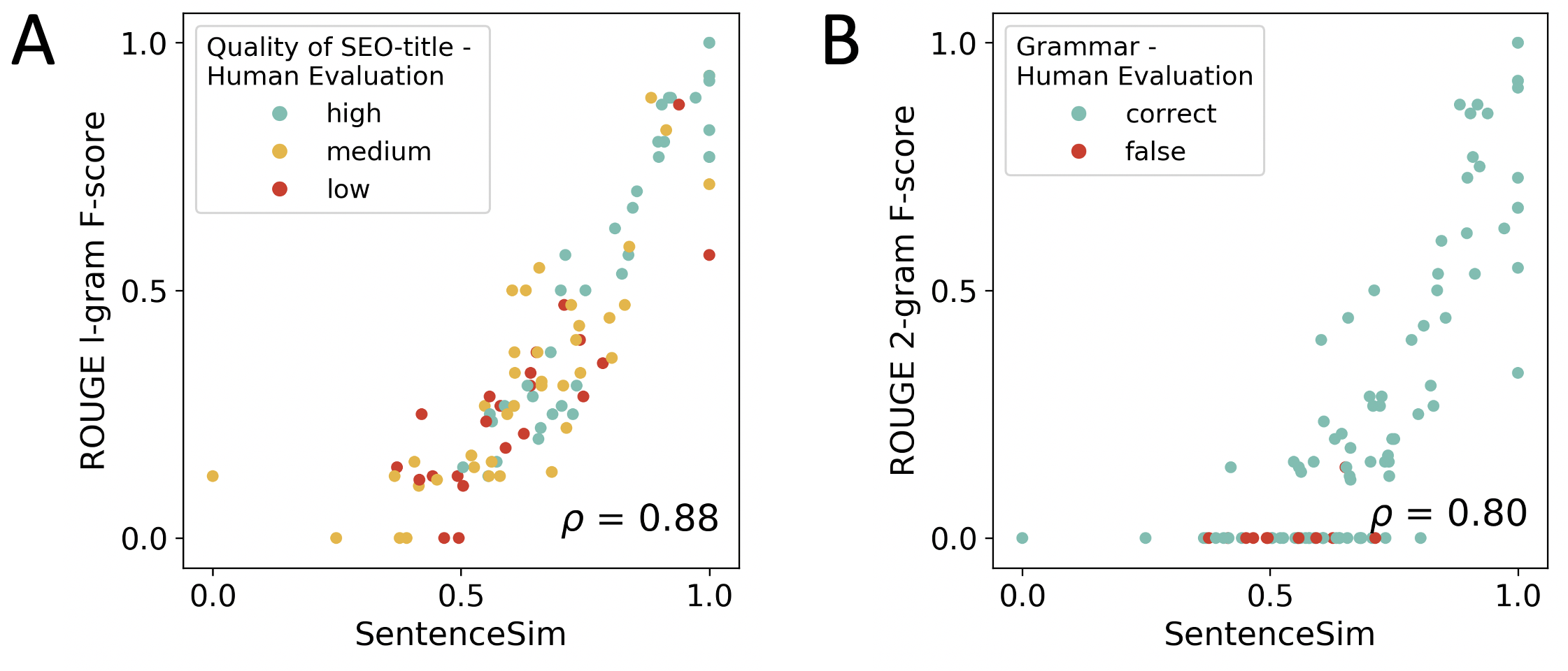}
     \caption{\textbf{The newly introduced SentenceSim correlates well with established metrics such as ROUGE l-gram F-score.} Displayed above are different metrics for the test set consisting of 100 news articles. We either automatically compare (ROUGE scores, SentenceSim) the generated SEO title with the already existing SEO title approved by SEO experts or use manual assessment of the articles. \textbf{A} For each generated title the ROUGE l-gram F-score against SentenceSim - our newly introduced metric - are shown. Both metrics have a correlation of $\rho=0.88$. The quality of the headline is color-coded with a quality score of 5 or 4.5 for "high quality", 4 or 3.5 for "medium quality" and 3 and below for "low quality". \textbf{B} For each generated title the ROUGE 2-gram F-score against SentenceSim are shown. Both metrics have a correlation of $\rho=0.80$. The grammar of the headline is color-coded with a score of 1 "correct grammar" and 0.5 to 0 for "false grammar". }
     \label{fig:metrics}
 \end{figure}

ROUGE scores \cite{lin-rouge} are an established method \cite{zhang2019pegasus,scialom-etal-2019-answers} for evaluating summarization. They measure the overlapping n-grams in the generated summary and the reference summary. We calculated for the aforementioned set of articles, ROUGE-1, ROUGE-2, and ROUGE-L, the most common metrics. Although the models have been trained and tested on different datasets, the results in Table \ref{tab:metrics} are slightly better than the results achieved with the English \textsc{BertSumAbs} model trained on a CNN-data set \cite{presumm}.
\begin{table}
\label{tab:metrics}
\begin{center}
\begin{tabular}{| c | c | c | } 
\hline
Score & German \textsc{BertSumAbs} & \thead{English \textsc{BertSumAbs}\\ \cite{presumm}} \\ \hline 
    ROUGE 1gram F-score  &	43.54 & 41.72 \\
    ROUGE 2gram F-score  &	24.84 & 19.39 \\	
    ROUGE lgram F-score  &	40.02 & 38.76 \\ [1ex] 
    SentenceSim &  68.19 & \\
    \hline\hline &&\\
    Grammar correct? & 92.42 &\\[0.5ex] 
    False information?      & 12.12 &\\  
    Informative?     & 3.88 &\\
    \hline
\end{tabular}
\end{center}
\caption{\textbf{Average results from the model evaluation}. The automated evaluation was done on 1000 articles. It includes calculation of ROUGE scores and is compared with result of then English \textsc{BertSumAbs}. All scores are percentages. SentenceSim is the newly introduced metric, cp. text. The manual evaluation was done on 100 articles. The first two questions were answered with either "yes"=1 or "no"=0. Display in percentage. For the last question we agreed on a discrete scale from 1=worst to 5=best, see Sec.\ref{sec:imp_details} for further details.  }
\end{table}
However, we find that ROUGE is not well suited for abstractive summarization as it measures the exact overlap of wording between generated title and reference. Valid summarizations using different words than the reference result in low ROUGE scores. Hence, we investigated a different approach based on word embeddings where words are mapped into a multi-dimensional vector space and high similarity between words semantics is reflected by high proximity within this vector space. To measure the similarity of whole sentences we introduced a metric which calculates the mean of maximum cosine similarity between words of the generated SEO title and its reference:
\begin{equation}
    S = \frac{1}{N}\sum_{i=1}^{N}\max_{1\leq j\leq M}\big(\text{sim}(x_i,y_j)\big),
\end{equation}
where $x_i$ is the vector representation of the $i$-th word in the generated summary with $N$ words, $y_j$ is the vector representation of the $j$-th word in the reference with $M$ words. sim is the cosine similarity. We will refer to this metric as SentenceSim.
The average of SentenceSim over all test samples shows a higher score compared to the ROUGE scores as expected and in this it can account better for different wordings. Additionally, there is a high correlation between SentenceSim and ROUGE-L ($rho = 0.88$) and also between SentenceSim and ROUGE-2 ($rho = 0.80$), cp. Fig. \ref{fig:metrics} which indicates that SentenceSim contains similar information.
 \subsection{Manual Evaluation}
 \label{sec:humaneval}
 While SentenceSim is able to give informative scores to valid SEO titles with different words than the reference, it has its limitations. It is not able to capture grammatical mistakes or misleading and false information. Therefore, we conducted a manual revision on a subset of 100 articles from the same data set based on 3 questions. Every generated headline was judged by 2 individuals and every individual had an overlap of her judgement with 2 others to handle bias in judgement. This human inspection revealed that the German \textsc{BertSumAbs} generates articles with mostly correct grammar and which rarely contain false information, cp. Table \ref{tab:metrics}. SEO titles with a high quality judged by humans have in most cases also a high SentenceSim score, cp. Fig. \ref{fig:metrics}A. Similarly, false grammar is characterized by low SentenceSim scores, cp. Fig.\ref{fig:metrics}B.  

\section{Summary and Outlook}
In this paper we introduce an approach to generate search engine optimized titles for German news articles. To achieve this, we modify \textsc{BertSumAbs} \cite{presumm}, an English language model for abstractive summarization:
We incorporate German Bert into the encoder and we alter the decoder so that it produces short titles by adjusting the length penalty introduced by \cite{opennmt}. We further present a rank penalty in order to favor titles containing keywords relevant for search engine optimization. 

\begin{figure}[h]
    \centering
    \includegraphics[width=1\textwidth]{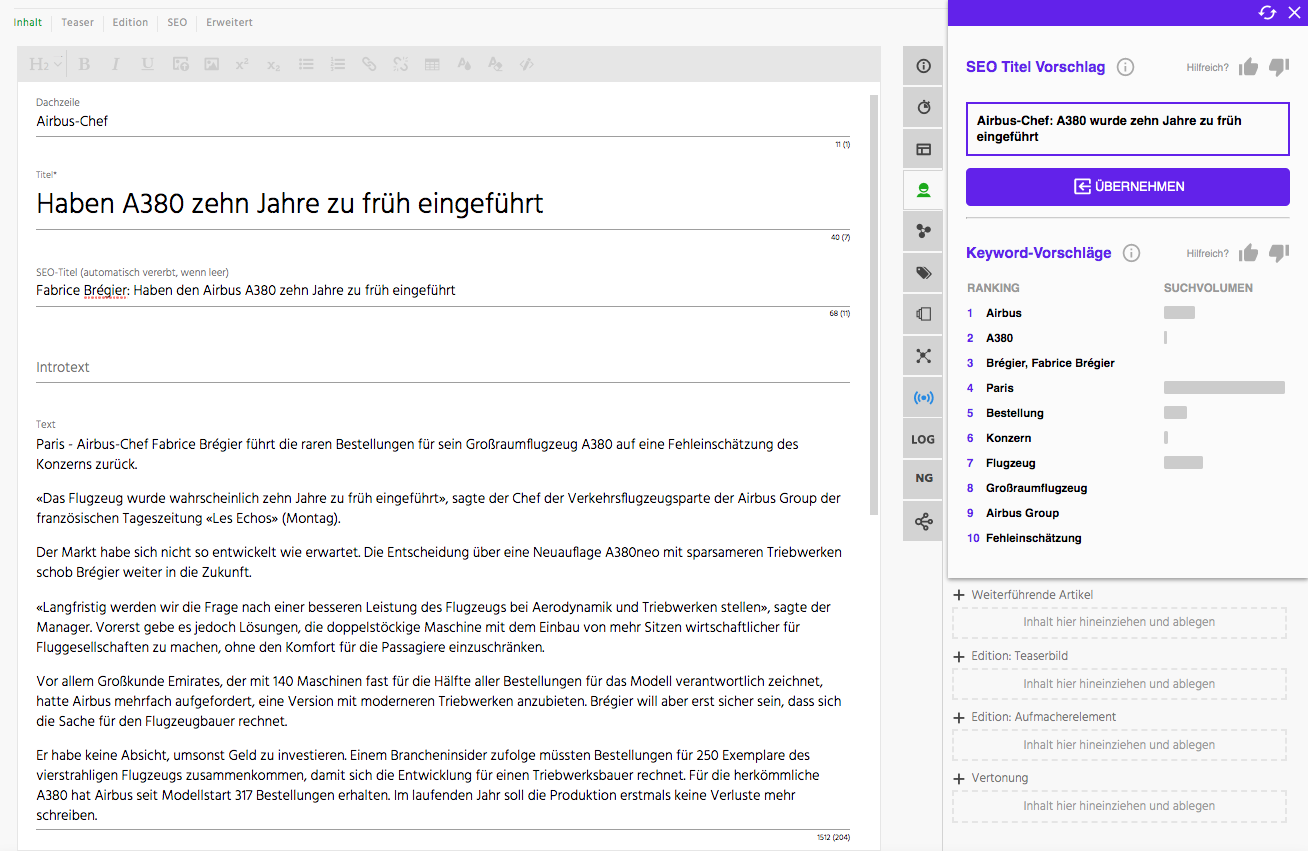}
    \caption{The browser plugin displays the title prediction and ranked keywords based on the article's text in the article editor. The generated title reads "Airbus boss: A380 was introduced ten years too early". The original SEO title in the editor reads: "Fabrice Br\'egier: We have introduced the Airbus A380 ten years too early". The example is taken from the test set used for manual evaluation.}
    \label{fig:screenshot}
\end{figure}

To make the title predictions accessible for the journalists, we developed a browser plugin that integrates with the article editor. The plugin displays both the title prediction and the list of ranked keywords in combination with the expected search volume, see \ref{fig:screenshot}. Every access to this browser plugin is logged. This will allow us not only to track the frequency of usage but also which articles have been influenced by the model outputs. After a user test phase, we will evaluate the collected data and compare the average search traffic generated by the test group with that of a control group.

Search traffic can vary strongly per article as it depends on the public interest - e.g. an article on local news can only gain little interest compared to articles about major topics such as the Corona-Virus. Since the underlying distribution is unknown, non-parametric statistics, especially permutation tests are suitable \cite{permutation}. A one-sided Monte Carlo significance test \cite{montecarlo} performed with enough permutations $N$ under the null hypothesis that both article groups come from the same distribution (i.e. generate the same mean traffic) would suffice to decide whether this null hypothesis can be rejected given a threshold p-value. In this way, we hope to see if deploying the model to production has a positive impact on search traffic arriving at the WELT website.

\section{Implementation Details}
\label{sec:imp_details}

We pre-process the data by splitting sentences with the Natural Language Toolkit \cite{stanford}. We implemented the model using Pytorch together with OpenNMT \cite{opennmt} and the 'bert-base-german-cased' \cite{germanbert} version of BERT made available by hugging face \cite{huggingface} to implement \textsc{BertSumAbs}. BERT's WordPiece tokenizer is used on both the source and the target text.
The code base that we build on is taken from \cite{presumm}, \cite{presumm_github}.

In our model, we apply dropout (probability of 0.1) before all linear layers and label smoothing with smoothing factor 0.1. The number of hidden units in the Transformer decoder is 768 and for all feed-forward layers, we have a hidden size of 2,048 \cite{presumm}. We use Adam optimizer for both the encoder and decoder. The encoder uses a learning rate of 0.002 and warmup steps of 20000 whereas the decoder uses a learning rate of 0.2 and warmup steps of 10000. The model was trained for 100,000 steps with batch size 140 on 4 GPUs (NVIDIA V100 Tensor Core).\footnote{Amazon Sagemaker ml.p3.8xlarge: https://aws.amazon.com/de/sagemaker/pricing/instance-types/} Gradient accumulation is applied every 5 steps. Every 2,000 steps we create model checkpoints, which are evaluated on a validation set of 10,000 articles. We select the best checkpoint based on evaluation accuracy and perplexity on the validation set.

In the decoder we use a beam search of size 10 and return the result of the best beam. Additionally, we use $\alpha=0.6$ for the length penalty and $\beta=1.5$ for the rank penalty. Reported bigrams, trigrams as well as repetitive words are blocked and we predict until an end-of-sequence token is emitted or the maximum length of 20 tokens is achieved. Because of the subword-tokenizer, we rarely observe out-of-vocabulary words in the output.

We use the Google entity recognition API \cite{google_ent} for extracting keywords from an article and pytrends \cite{pytrends} for gathering their comparing relative search volumes, fed as inputs to the XGBoost ranking algorithm.

We report average result on the test set of 1000 data points for automated evaluation. Additionally, we check out of those 100 data points manually. The automated evaluation included ROUGE and our newly introduced SentenceSim.
The ROUGE-metrics were calculated using Py-rouge \cite{pyrouge}. For the word embedding used in SentenceSim we used the German FastText embedding \cite{fasttext} together with the Gensim API \cite{gensim}. For the manual assessment we used 3 different questions together with a scale. The questions included both binary and discrete ones. The binary questions, i.e. "Is grammar correct?" and "Does the title contain false information?", were answered with 1 = "yes" and 0 = "no". To examine the quality of the title we agreed on a discrete scale: 1 = "Not informative, no meaning", 2 = "Only selective parts of the content covered, gives a wrong impression of what the article is about", 3 = "gives you a very brief but correct impression of what the article is about", 4 = "all important information included, you know what you have to expect", 5 = "all important information and a good headline (nice structure, makes reader curious)". Every title was judged by two individuals, each having an overlap with two others. The final score for each title was calculated by the mean of both judgements. 

%%%%%%%%%%%%%%%%%%%%%%%%%%%%%%%%%%%%%%%%%%%%%%%%%%%%%%%%%%%%%%%%%%%%%%
\begin{acknowledgment}
We thank the SEO team and the WELT editorial team for fruitful discussions and their patience with answering our endless questions.
\end{acknowledgment}

%%%%%%%%%%%%%%%%%%%%%%%%%%%%%%%%%%%%%%%%%%%%%%%%%%%%%%%%%%%%%%%%%%%%%%
% The bibliography is stored in an external database file
% in the BibTeX format (file_name.bib).  The bibliography is
% created by the following command and it will appear in this
% position in the document. You may, of course, create your
% own bibliography by using thebibliography environment as in
%
% \begin{thebibliography}{12}
% ...
% \bibitem{itemreference} D. E. Knudsen.
% {\em 1966 World Bnus Almanac.}
% {Permafrost Press, Novosibirsk.}
% ...
% \end{thebibliography}

% Here's where you specify the bibliography style file.
% The full file name for the bibliography style file 
% used for an ASME paper is asmems4.bst.
\bibliographystyle{asmems4}

% Here's where you specify the bibliography database file.
% The full file name of the bibliography database for this
% article is asme2e.bib. The name for your database is up
% to you.
\bibliography{asme2e}

%%%%%%%%%%%%%%%%%%%%%%%%%%%%%%%%%%%%%%%%%%%%%%%%%%%%%%%%%%%%%%%%%%%%%%

\end{document}